\documentclass[11pt,a4paper]{article}
\usepackage[hyperref]{naaclhlt2019}
\usepackage{times}
\usepackage{latexsym}
\usepackage{algorithm, algpseudocode}
\algnewcommand{\LeftComment}[1]{\State \(\triangleright\)#1}
\usepackage{tabularx}
\usepackage{float}
\usepackage{enumitem}
\usepackage{siunitx}
\usepackage{soul}

\usepackage{url}
\aclfinalcopy 
\def\aclpaperid{1790} 

\usepackage{todonotes}

\newcommand\Mark[1]{\textsuperscript#1}

\title{Curriculum Learning for Domain Adaptation\\in Neural Machine Translation}

\author{Xuan Zhang\Mark{1}, Pamela Shapiro\Mark{1}, 
Gaurav Kumar\Mark{1}, Paul McNamee\Mark{1},\\ 
{\bf Marine Carpuat\Mark{2}}, 
{\bf Kevin Duh\Mark{1}}\\[1em]
\Mark{1}Johns Hopkins University\\
\Mark{2}University of Maryland\\[1em]
\texttt{\{xuanzhang, pshapiro, mcnamee\}@jhu.edu,}\\
\texttt{marine@cs.umd.edu, \{gkumar, kevinduh\}@cs.jhu.edu}\\[1em]
}

\date{}

\begin{document}
\maketitle

\begin{abstract}
We introduce a curriculum learning approach to adapt generic neural machine translation models to a specific domain. Samples are grouped by their similarities to the domain of interest and each group is fed to the training algorithm with a particular schedule. This approach is simple to implement on top of any neural framework or architecture, and consistently outperforms both unadapted and adapted baselines in experiments with two distinct domains and two language pairs.

\end{abstract}

\section{Introduction}
\label{sec:introduction}
Neural machine translation (NMT) performance often drops when training and test domains do not match and  when in-domain training data is scarce \citep{koehn2017six}. Tailoring the NMT system to each domain could improve performance, but unfortunately high-quality parallel data does not exist for all domains. 
Domain adaptation techniques address this problem by exploiting diverse data sources to improve in-domain translation, including {\em general domain} data that does not match the domain of interest, and {\em unlabeled domain} data whose domain is unknown (e.g.~webcrawl like Paracrawl).

\begin{figure}[t]
    \centering
    \includegraphics[width=\linewidth]{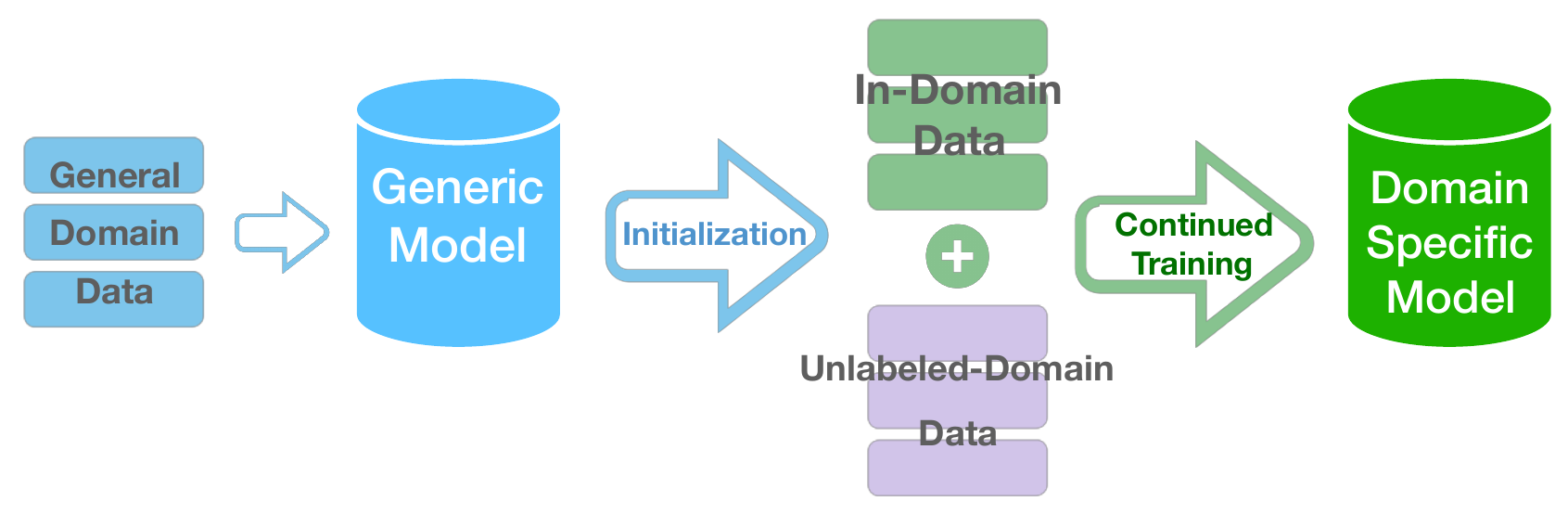}
    \caption{Workflow of our domain adaptation system.}
    \label{fig:intro}
\end{figure}

One approach to exploit unlabeled-domain bitext is to apply {\em data selection} techniques \citep{moore2010intelligent, axelrod2011domain, duh2013adaptation} to find bitext that are similar to in-domain data. This selected data can additionally be combined with in-domain bitext and trained in a continued training framework, as shown in Figure \ref{fig:intro}. Continued training or fine-tuning \citep{luong2015effective, freitag2016fast, chu2017empirical} is an adaptation technique where a model is first trained on the large general domain data, then used as initialization of a new model which is further trained on in-domain bitext. In our framework, the selected samples are concatenated with in-domain data, then used for continued training. This effectively increases the in-domain training size with ``pseudo" in-domain samples, and is helpful in continued training \citep{koehn-duh-thompson:2018:WMT}.

A challenge with employing data selection in continued training is that there exists no clear-cut way to define whether a sample is sufficiently similar to in-domain data to be included. In practice, one has to define a threshold based on similarity scores, and even so the continued training algorithm may be faced with samples of diverse similarities. We introduce a new domain adaptation technique that addresses this challenge.

Inspired by curriculum learning \citep{bengio2009curriculum}, we use the similarity scores given by data selection to rearrange the order of training samples, such that {\em more similar examples are seen earlier and more frequently during training}.
To the best of our knowledge, this is the first work applying curriculum learning to domain adaptation.

We demonstrate the effectiveness of our approach on TED Talks and patent abstracts for German-English and Russian-English pairs, 
using two distinct data selection methods, Moore-Lewis method \citep{moore2010intelligent} and cynical data selection \citep{Axelrod:2017}.
Results show that our approach consistently outperforms standard continued training, with up to 3.22 BLEU improvement. Our S$^4$ error analysis \citep{irvine2013measuring} reveal that this approach reduces a reasonable number of SENSE and SCORE errors.


\section{Curriculum Learning for Adaptation}
\label{sec:cl4da}
\citet{weinshall2018curriculum} provide guidelines for curriculum learning: ``A practical curriculum learning method should address two main questions: how to rank the training examples, and how to modify the sampling procedure based on this ranking." For domain adaptation we choose to estimate the difficulty of a training sample based on its distance to the in-domain data, which can be quantified by existing data selection methods (Section \ref{ssec:dselection}). 
For the sampling procedure, we adopt a probabilistic curriculum training (CL) strategy that takes advantage of the spirit of curriculum learning in a nondeterministic fashion without discarding the good practice of original standard training policy, like bucketing and mini-batching. 

\subsection{Domain Similarity Scoring}
\label{ssec:dselection}
We adopt similarity metrics from prior work on data selection to score examples for curriculum learning. Let $I$ be an in-domain corpus, and $N$ be a unlabeled-domain data set. Data selection models rank sentences in $N$ according to a domain similarity measure with respect to $I$, and choose top $n$ samples from $N$ by a cut-off threshold for further training purpose. We examine two data selection methods, Moore-Lewis method \cite{moore2010intelligent} and cynical data selection \cite{Axelrod:2017}.\\

\noindent{\bf Moore-Lewis Method} 
Each sentence $s$ in $N$ is assigned a {\em cross-entropy difference score},
\begin{equation}
    H_I(s) - H_N(s),
\end{equation}
where $H_I(s)$ is the per-word cross-entropy of $s$ according to a language model trained on $I$, and
$H_N(s)$ is the per-word cross-entropy of $s$ according to a language model trained on a random sample of $N$ with roughly the same size as $I$. 
A lower cross-entropy difference indicates that $s$ is more like the in-domain data and less like the unlabeled-domain data.\\

\noindent {\bf Cynical Data Selection} Iteratively select sentence $s$ from $N$ to construct a training corpus that would approximately model $I$. At each iteration, each sentence is scored by the expected cross-entropy change from adding it to the already selected subset of $N$. The selected sentence is the one which most decreases $H_n$, the cross-entropy between previously selected $n$-sentence corpus and $I$.

\subsection{Curriculum Learning Training Strategy}
\label{ssec: tstrategy}

We identify two general types of curriculum learning strategy. The {\it deterministic} curriculum (c.f.~\citet{kocmi2017curriculum})  trains on a fixed order of samples based on their scores (e.g.~``easy-to-hard" or ``more similar to less"). While simple to motivate, this may not always perform well because neural methods benefit from randomization in the minibatches and multiple epochs. In contrast, the {\it probabilistic} curriculum \cite{bengio2009curriculum} works by dividing the training procedure into distinct phases. Each phase creates a random sample from the entire pool of data, but earlier phases sample the ``easier" or ``more similar" sentence with higher probability.. Since each phase can be viewed as creating a new training dataset, all the well-tested tricks of the trade for neural network optimization can be employed.

In this paper, we use the same probabilistic curriculum strategy and code base\footnote{\url{https://github.com/kevinduh/sockeye-recipes/tree/master/egs/curriculum}} as \citet{zhang2018empirical}. The main difference here is the application to domain adaptation. The proposed strategy is summarized as follows:
\begin{itemize}[noitemsep,topsep=0pt,parsep=0pt,partopsep=0pt]
    \item Sentences are first ranked by similarity scores and then distributed evenly into shards, such that each shard contains samples with similar similarity criteria values.
    \item The training process is segmented into consecutive {\em phases}, where only a subset of shards are available for training. 
    \item During the first phase, only the easiest shard is presented. When moving to the next phase, the training set will be increased by adding the second easiest shard into it, and so on. Easy shards are those that are more similar to the in-domain data, as quantified by either Moore-Lewis or Cynical Data Selection.
    \item The presentation order of samples is not deterministic. (1) Shards within one curriculum phase are shuffled, so they are not necessarily visited by the order of similarity level during this phase. (2) Samples within one shard are bucketed by length and batches are drawn randomly from buckets. 
\end{itemize}

\section{Experiments and Results}
\label{sec:e&r}
We evaluate on four domain adaptation tasks. The code base is provided to ensure reproducibility.\footnote{\url{https://github.com/kevinduh/sockeye-recipes/tree/master/egs/curriculum}}
\subsection{Data and Setup}
\label{ssec:data}
\noindent{\bf General Domain Data} 
We have two general domain datasets, Russian-English (ru) and German-English (de). Both are a concatenation of 
OpenSubtitles2018 \cite{lison2016opensubtitles2016}
and WMT 2017 \citep{bojar2017findings},
which contains data from several domains, e.g.~parliamentary proceedings (Europarl, UN Parallel Corpus), political/economic news (news commentary, Rapid corpus), and web-crawled parallel corpus (Common Crawl, Yandex, Wikipedia titles). We performed sentence length filtering (up to 80 words) after tokenization, ending up with 28 million sentence pairs for German and 51 million sentence pairs for Russian.
\\

\noindent {\bf In-domain Data} We evaluate our proposed methods on two distinct domains per language pair: 
\begin{itemize}[noitemsep,topsep=0pt,parsep=0pt,partopsep=0pt]
    \item TED talks: data-split from \citet{duh18multitarget}.
    \item Patents:~from the World International Property Organization COPPA-V2 dataset \citep{JunczysDowmunt2016COPPAV2}. 
\end{itemize}
We randomly sample 15k parallel sentences from the original corpora as our in-domain bitext.\footnote{Appendix \ref{app:indata} explains our choice of 15k in detail.} We also have around 2k sentences of development and test data for TED and 3k for patent.\\

\noindent{\bf Unlabeled-domain Data} For additional unlabeled-domain data, we use web-crawled bitext from the Paracrawl project.\footnote{\url{https://www.paracrawl.eu/}} We filter the data using the Zipporah cleaning tool~\cite{D17-1319}, with a threshold score of 1.  
After filtering, we have around 13.6 million Paracrawl sentences available for German-English and 3.7 million Paracrawl sentences available for Russian-English. Using different data selection methods, we include up to the 4096k and 2048k sentence-pairs for our German and Russian experiments, respectively.
\\


\noindent{\bf Data Preprocessing} All datasets are tokenized using the Moses \cite{koehn2007moses} tokenizer. We learn byte pair encoding (BPE) segmentation models \cite{sennrich2015neural} from general domain data. The BPE models are trained separately for each language, and the number of BPE symbols is set to 30k. We then apply the BPE models to in-domain and Paracrawl data, so that the parameters of the generic model can be applied as an initialization for continued training. Once we have a converged generic NMT model, which is very expensive to train, we can adapt it to different domains, without building up a new vocabulary and retraining the model. \\
\begin{figure*}
    \centering
    \includegraphics[width=\linewidth]{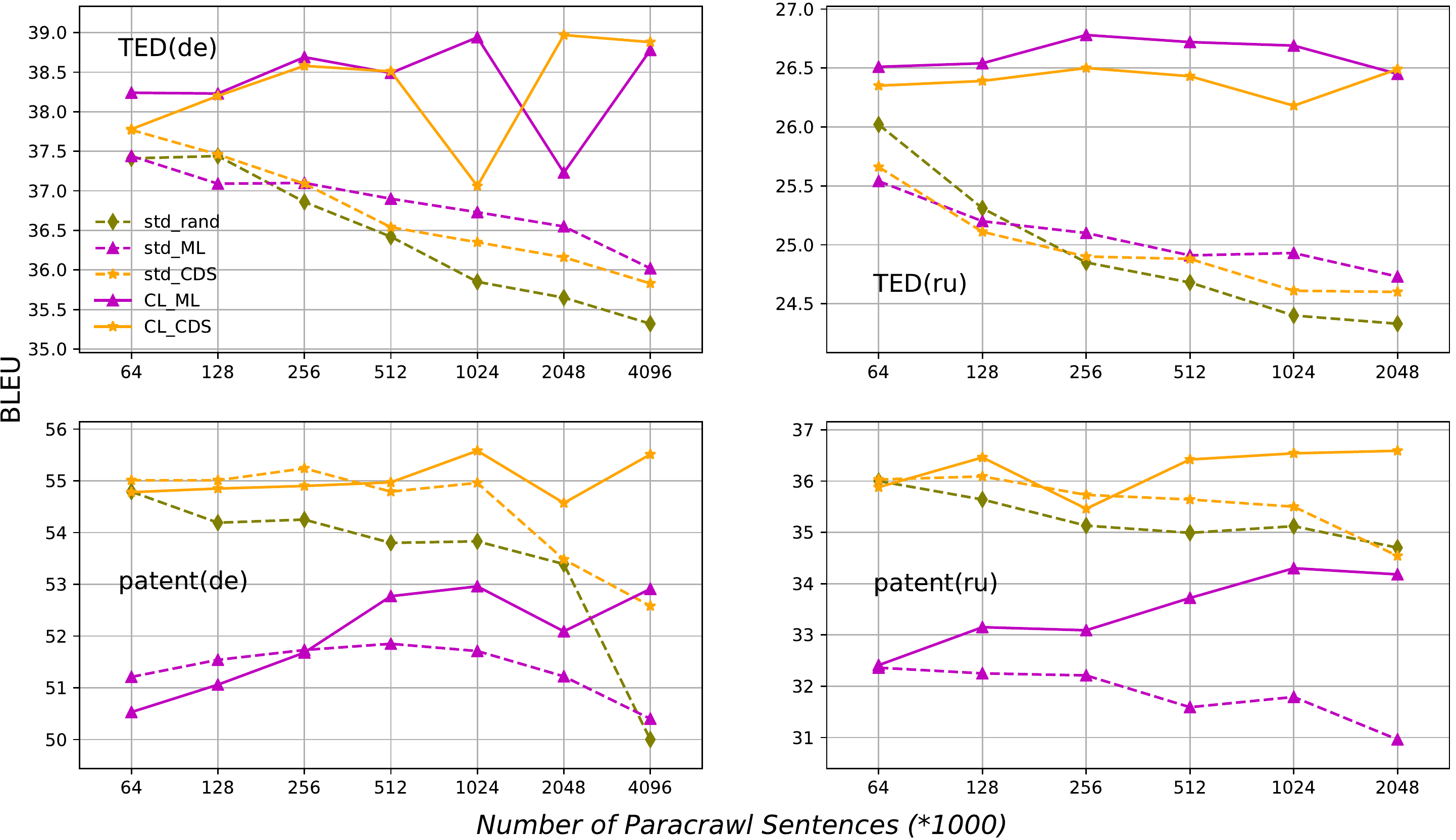}
    \caption{BLEU of adapted models using a concatenation of in-domain and varying amounts of Paracrawl data.
    }
    \label{fig:results}
\end{figure*}

\noindent{\bf NMT Setup} Our NMT models are developed in Sockeye\footnote{\url{github.com/awslabs/sockeye}} \cite{hieber2017sockeye}. 
The generic model and continued training model are trained with the same hyperparameters. We use the seq2seq attention architecture \citep{bahdanau2014neural} with 2 LSTM layers for both encoder and decoder, and 512 hidden nodes in each layer. The word embedding size is also set to 512. Our models apply Adam \cite{kingma2014adam} as the optimizer, with an initial learning rate 0.0003. 
The learning rate is multiplied by 0.7 whenever validation perplexity does not surpass the previous best in 8 checkpoints.\footnote{The Adam optimizer for continued training model is initialized without reloading from the trained generic model.}  We use minibatches of 4096 words. Training stops when the perplexity on the development set has not improved for 20 checkpoints (1000 updates/batches per checkpoint). \\



\noindent{\bf Domain Similarity Scoring Setup} To get similarity scores, we build 5-gram language models on the source side\footnote{Appendix \ref{app:bilingual} also shows the effect of using language models built from target side and both sides.} with modified Kneser-Ney smoothing using KenLM \citep{heafield2011kenlm}.\\

\noindent{\bf Curriculum Learning Setup} 
The number of batches in each curriculum phase is set to 1000. We split the training data into 40 shards\footnote{After experimenting with various values from 5 to 100 (Appendix \ref{app:dsharding}), we found best performance can be achieved at 40 shards.}, with all the 15k in-domain data in the first shard, and Paracrawl data split into the remaining 39 shards. 

\subsection{Experimental Comparison}
\label{ssec:esetup}
Our goal is to empirically test whether the proposed curriculum learning method improves translation quality in the continued training setup of Figure \ref{fig:intro}. 
We compare two approaches to continued training: (1) the standard approach reads batches of in-domain and selected Paracrawl in random order; (2) the proposed curriculum learning approach reads these batches according to a schedule. We run the comparison with two data selection methods, leading to four systems: 
\begin{itemize}[noitemsep,topsep=0pt,parsep=0pt,partopsep=0pt]

    \item {\bf std\_ML}: standard continued training with Moore-Lewis scores
    \item {\bf CL\_ML}: curriculum learning approach to continued training with Moore-Lewis scores
    \item {\bf std\_CDS}: standard continued training with scores from Cynical Data Selection
    \item {\bf CL\_CDS}: curriculum learning approach to continued training with scores from Cynical Data Selection
\end{itemize}

For reference, we show results of the generic model ({\bf GEN}), the model trained from scratch with in-domain data ({\bf IN}), the model continued trained on in-domain data only ({\bf IN\_CT}), and a standard continued training model using a random subset (rather than ML or CDS scores) of the concatenated in-domain and Paracrawl data ({\bf std\_rand}).

\subsection{Results}
\label{ssec:results}
\begin{table}
    \centering
    \small
    \begin{tabular}{p{0.85cm} S[table-format=2.2] S[table-format=2.2] S[table-format=2.2] S[table-format=2.2]}
       & {\bf TED(de)}  &  {\bf TED(ru)} & {\bf patent(de)} & {\bf patent(ru)}\\
       \hline
       {\bf GEN} & 34.59 & 23.40 & 35.95 & 23.41 \\
       {\bf IN} & 2.53 & 1.76 & 12.09 & 16.81 \\
       {\bf IN\_CT} & 36.16 & 25.04 & 54.70 & 35.61\\
       {\bf std\_rand} & 35.32 & 24.33 & 50.00 & 34.70 \\
       \hline
       {\bf std\_ML} & 36.02 & 24.73 & 50.40 & 30.96 \\
       {\bf CL\_ML} &  38.78 &  26.45 & 52.91 & 34.18 \\
       {\bf $\Delta$\_ML}& +2.76 & +1.72 & +2.51 & +3.22\\
       \hline
       {\bf std\_CDS} & 35.83 & 24.60 & 52.58 & 34.54\\
       {\bf CL\_CDS} & {\bf 38.88} & {\bf 26.49} & {\bf 55.51} & {\bf 36.59} \\
       {\bf $\Delta$\_CDS} & +3.05 & +1.89 & +2.93 & +2.05\\
      \hline
       
    \end{tabular}
    \caption{BLEU of unadapted \& adapted models. $\Delta$ shows improvement of {\bf CL} over {\bf std}.}
    \label{tab:results}
\end{table}
Table~\ref{tab:results} summarizes the key results, where we continue train on 15k in-domain samples and 4096k Paracrawl samples (for de) or 2048k Paracrawl samples (for ru):
\begin{itemize}[noitemsep,topsep=0pt,parsep=0pt,partopsep=0pt]
    \item The baseline BLEU scores confirm the need for domain adaptation. Using only the 15k in-domain samples alone (IN) is not sufficient to train a strong domain specific model, yielding BLEU scores as low as 2.53 on TED(de) and 1.76 on TED(ru). The model trained with a large amount of general domain data (GEN) is a stronger baseline, with BLEU scores of 34.59 and 23.40.
    \item Standard continued training is not robust to samples that are noisy and less similar to in-domain. As expected, continued training on in-domain data (IN\_CT) improves BLEU significantly, by up to 18.74 BLEU on patent(de). However, when adding  Paracrawl data, the standard continued training strategy (std\_rand, std\_ML, std\_CDS) consistently performs worse than IN\_CT. 
    \item Curriculum learning consistently improves BLEU score. Ranking examples using Moore-Lewis (CL\_ML) and Cynical Data Selection (CL\_CDS) improve BLEU over their baselines (std\_ML and std\_CDS) by up to 3.22 BLEU points.
\end{itemize}

As an additional experiment, we report results on different amounts of Paracrawl data.
Figure \ref{fig:results} shows how the curriculum uses increasing amounts of Paracrawl better than standard continued training. Standard continued training model hurts BLEU when too much Paracrawl data is added: for TED(de), there's a 1.94 BLEU drop when increasing CDS data from 64k to 4096k, and for patent(de), the decrease is 2.43 BLEU. By contrast, the curriculum learning models achieve a BLEU score that is as good or better as the initial model, even after being trained on the most dissimilar examples. This trend is clearest on the patent(ru) CL\_ML model, where the BLEU score consistently rises from 32.41 to 34.18.

The method used to score domain relevance has a different impact on the TED domain (top plots) and on the patent domain (bottom plots). On the patent domain, which is more distant from Paracrawl, CDS significantly outperforms ML. Replacing ML with CDS improve BLEU from 2.18 to 4.05 BLEU points for standard models and 2.20 to 4.25 BLEU points for curriculum learning models. Interestingly, for patents, the Moore-Lewis method does not beat the random selection, even when curriculum learning is applied. For example, at 64k selected sentences for patent(de), std\_rand gets 4.26 higher BLEU scores than CL\_ML. By contrast on the TED domain, which is closer to Paracrawl, the Moore-Lewis method slightly outperforms cynical data selection.
Due to these differences, we suggest trying different data selection methods with curriculum learning on new tasks; a potential direction for future work may be a curriculum that considers multiple similarity scores jointly. 

\section{Analysis}
\label{sec:analysis}

\subsection{Comparison of Curriculum Strategies}
\label{ssec:comparison}
We compare our approach to other curriculum strategies. {\em CL\_reverse} reverses the presenting order of the shards, so that shards containing less similar examples will be visited first, {\em CL\_scrambled} is a model that adopts the same training schedule as {\em CL}, but no data selection method and ranking is involved here --- Paracrawl data are evenly split and randomly assigned to shards; {\em CL\_noshuffle} is another curriculum learning model that does not shuffle shards in each curriculum phase.

\begin{figure}
    \centering
    \includegraphics[width=\linewidth]{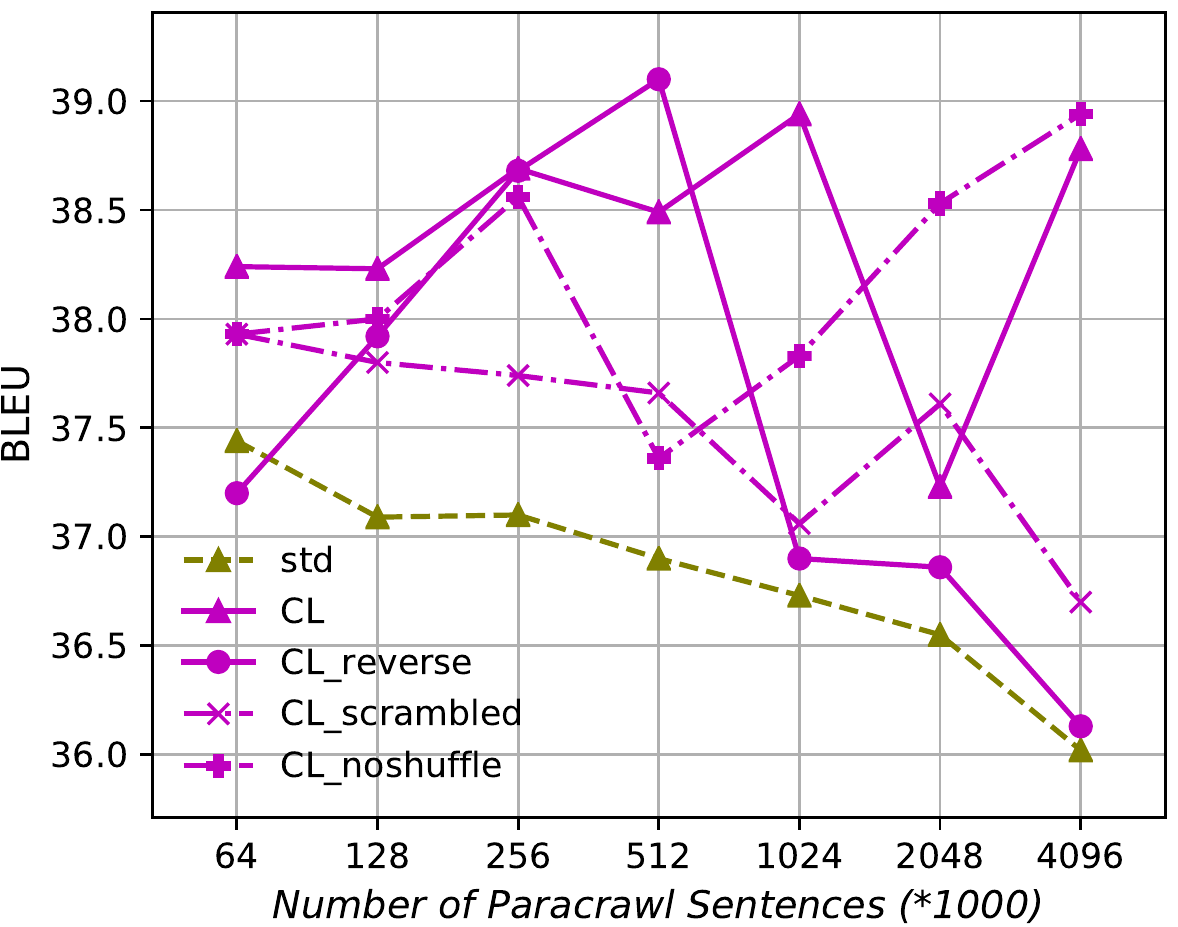}
    \caption{Comparison of various curriculum strategies on German-English TED corpora, where Moore-Lewis method is applied. \footnotemark}
    \label{fig:compare}
\end{figure}
\footnotetext{Each point represents a model trained to convergence on the fixed amount of in-domain and ParaCrawl data whose amount is specified by the x-axis.}

Results from Figure \ref{fig:compare} show that CL outperforms CL\_reverse and CL\_noshuffle for 5 out of 7 cases and outperforms CL\_scrambled in 6 out of 7 cases. This suggests that it is beneficial to train on examples that are closest to in-domain first and to use a probabilistic curriculum. Analyzing the detailed difference between CL and CL\_reverse would be interesting future work.
One potential hypothesis why CL might help is that it first trains on a low-entropy subset of the data before moving on to the whole training set, which may have regularization effects.
\begin{figure}[h]
    \centering
    \includegraphics[width=\linewidth]{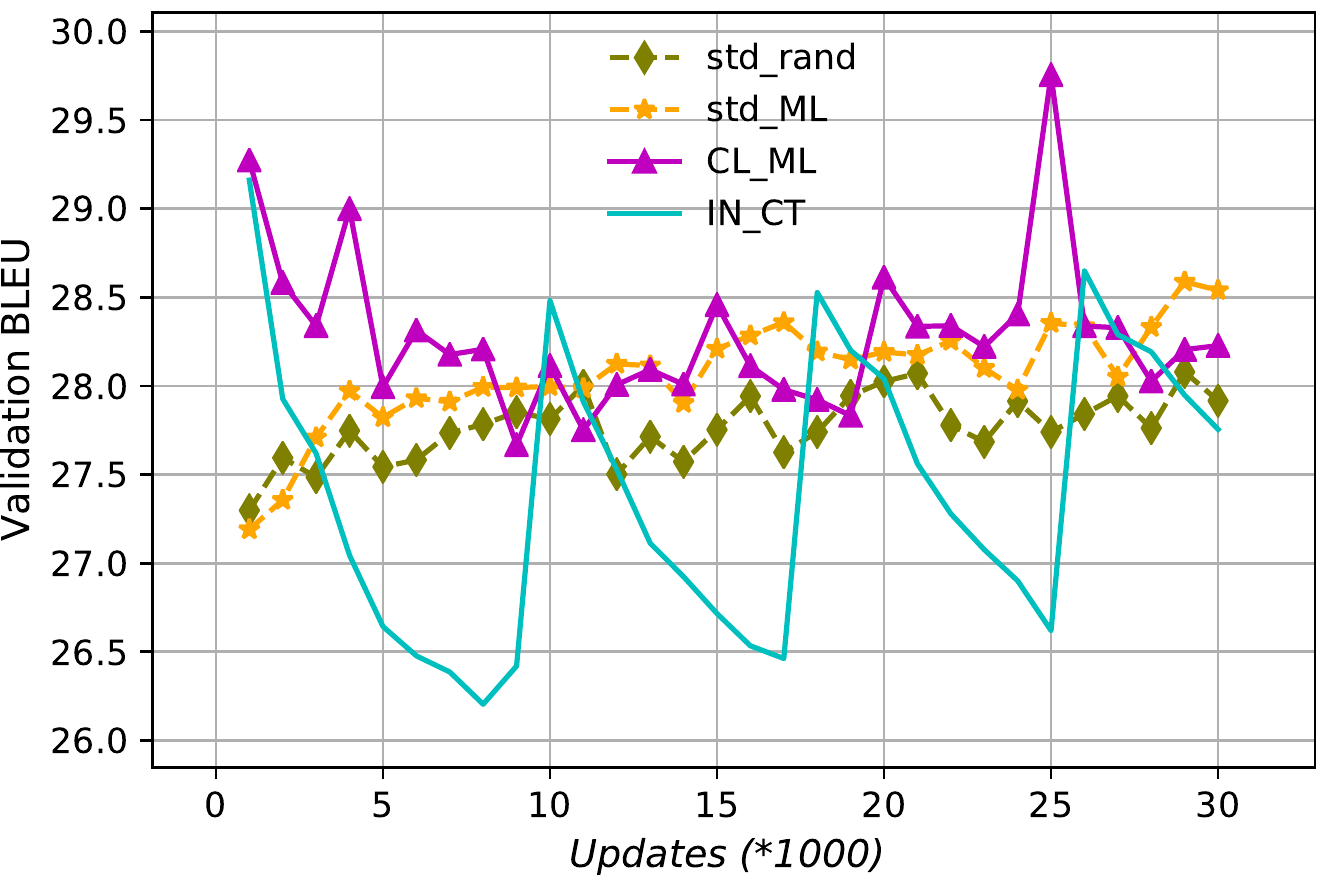}
    \caption{Learning curves for German-English TED NMTs. Except for IN\_CT, the other three models all continued trained on the concatenation of in-domain and 1024k Paracrawl data.}
    \label{fig:lcurves}
\end{figure}

\subsection{Learning Curves}
\label{ssec:lcurves}

Learning curves (Figure {\ref{fig:lcurves}}) further illustrate the advantage of our method. Continued training on in-domain data only starts from a strong initialization (thanks to pre-training on large general domain data) but heavily oscillates over training without reaching the initial performance. 
This behavior may be due to the sparsity of the TED data: the small randomly sampled training set may not represent the development and test data well.  Std\_ML shows opposite behavior to IN\_CT: it starts from a lower initial performance, and then gradually improves to a level comparable to IN\_CT. Std\_rand behaves similarly to std\_ML---in other words, uniformly sampling from Paracrawl drags the initial performance down without helping with the final performance.



Compared to all above, the curriculum learning models start from a high initial performance, suffer much less oscillation than IN\_CT, and gradually achieve the highest performance.\footnote{When converged, IN\_CT does not outperform CL\_ML.}

\begin{figure*}[t]
    \centering
    \includegraphics[width=\linewidth]{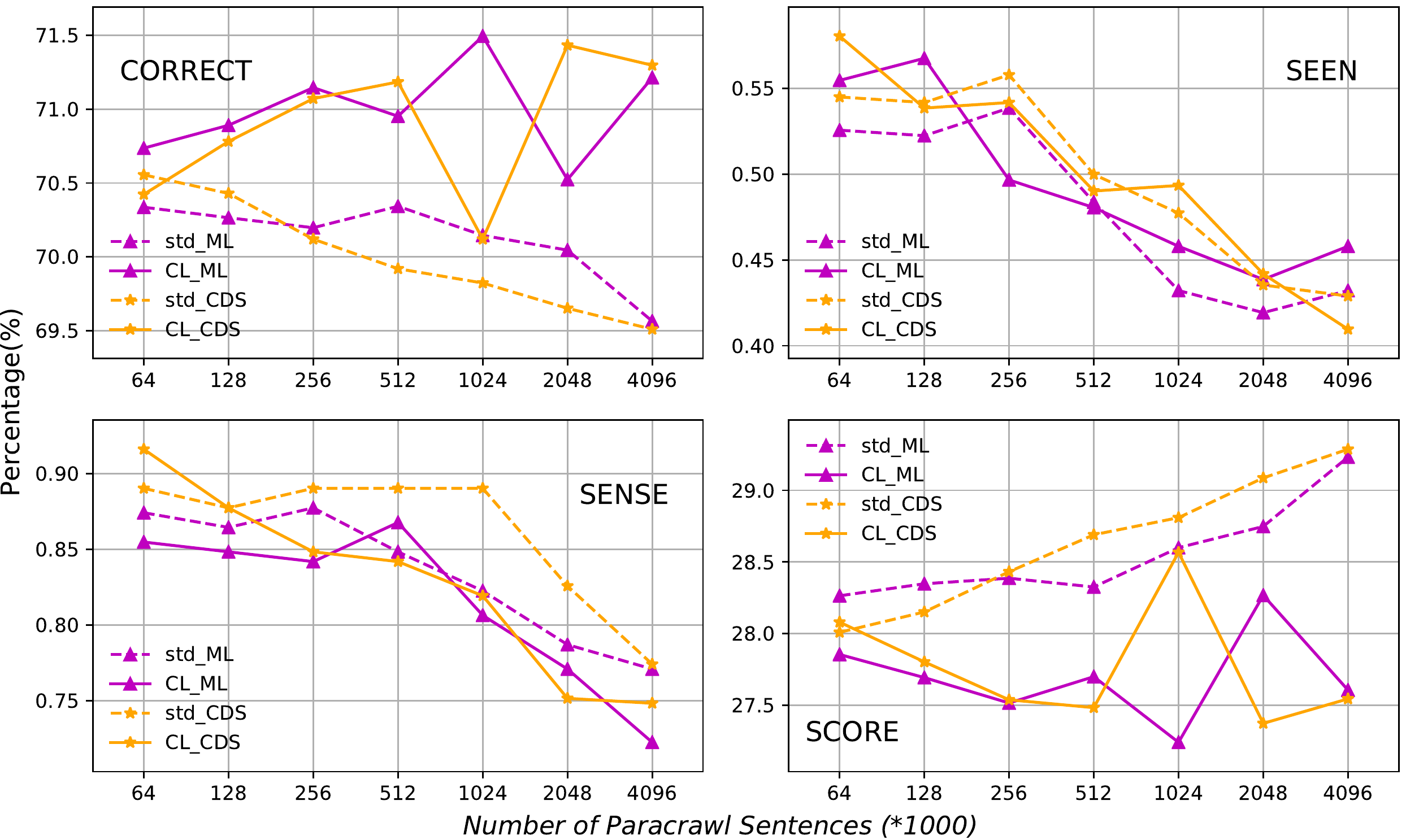}
    \caption{S$^4$ error analysis on German-English TED.}
    \label{fig:s4}
\end{figure*}
\subsection{Impact of Curriculum Learning on Lexical Choice: S$^4$ Analysis}
\label{ssec:s4}
How do translations improve when using curriculum learning? We characterize the impact of curriculum learning on lexical translation errors using the S$^4$ taxonomy of domain change errors introduced by \citet{irvine2013measuring} for phrase-based machine translation: (1) SEEN: incorrect translation for a source word that has never been seen in the training corpus; (2) SENSE: incorrect translation for a previously seen source word, whose correct translation (sense) has never been seen in the training corpus; (3) SCORE: a score error is made when the source word and its correct translation are both observed in training data, but the incorrect translation is scored higher than the correct alternative ; and (4) SEARCH: an error caused by pruning in beam search\footnote{We will only focus on the first three error categories in this paper for the purpose of model comparison.}. 

We extend this taxonomy to neural machine translation. As the unit of S$^4$ analysis is word alignment between a source word and a reference target word, we first run fast-align \cite{dyer2013simple} to get the source-target word alignments. After this, we follow the algorithm shown in Appendix \ref{app:s4} to give a summary of S$^4$ errors on the model's translation of test set.  

Figure \ref{fig:s4} shows the word translation results for the test set of German-English TED. Most of the errors are SCORE errors, while SEEN and SENSE errors are relatively rare. Curriculum learning significantly improves the adapted NMT systems at the word level --- with 4096k Paracrawl data selected by CDS, curriculum continued training model can translate 554 more words correctly than the standard continued training model. This improvement mainly happens in SCORE errors: 1.75\% of SCORE errors are corrected. SEEN and SENSE errors are also reduced by 0.02\% and 0.026\%, respectively. But overall, CL does not help much on SEEN errors.


\subsection{Characteristics of Selected Data}
\label{ssec:dsmethods}
We characterize the sentences chosen by different data selection methods, to understand their effect on adaptation as observed in Section~\ref{ssec:results}.\\

\begin{table*}
    \centering
    \begin{tabularx}{\textwidth}{l X}
    \hline
    {\bf TED\_ML} & It changes the way we think; it changes the way we walk in the world; it changes our responses; it changes our attitudes towards our current situations; it changes the way we dress; it changes the way we do things; it changes the way we interact with people.\\
    \hline
    {\bf TED\_CDS} & But, on the other hand, this signifies that the right of self-determination, as a part of the proletarian peace program, possesses not a ``Utopian'' but a revolutionary character.\\ 
    \hline
    {\bf patent\_ML} & The sites x, y and z can accommodate a large variety of cations with x=na+, k+, ca2+, vacancy; y=mg2+, fe2+, al3+, fe3+, li+, mn2+ and z=al3+, mg2+ , fe3+, v3+, cr3+; while the t site is predominantly occupied by si4+.\\
    \hline 
    {\bf patent\_CDS} & To select alternative viewing methods, such as for 3d-tv.\\
    \hline
    \end{tabularx}
    \caption{The top ranked sentences selected from German-English Paracrawl corpus.}
    \label{tab:sentexample}
\end{table*}
\begin{figure}
    \centering
    \includegraphics[width=\linewidth]{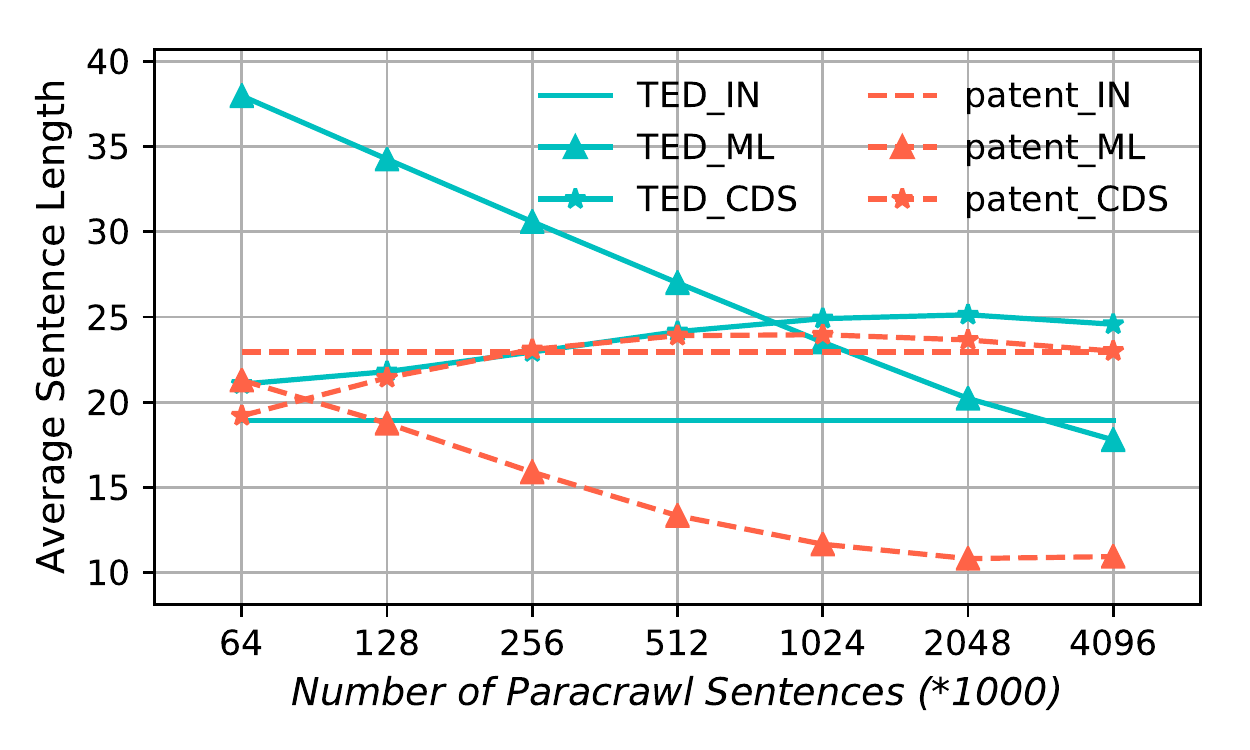}
    \caption{Average sentence length for increasing size of Paracrawl data. This is calculated on the source side of German-English pairs. TED\_IN stands for TED corpus. TED\_ML and TED\_CDS represent the Paracrawl samples selected by ML and CDS methods.}
    \label{fig:length}
\end{figure}

\noindent{\bf Selected Sentences Overlap} For each domain in German-English, we compute the overlap between the top $n$ ML and CDS Paracrawl sentences. The overlap is as low as 3.69\% for the top 64k sentences in the TED domain, and 8.43\% for the patent domain. Even in the top 4096k sentences, there are still 46.25\% and 65.40\% different ones in TED and patent domain respectively. See Table \ref{tab:sentexample} for examples of selected sentences.\\

\noindent {\bf Average Sentence Length} 
The ML score prefers longer sentences and is more correlated with sentence length (See Figure \ref{fig:length}) --- the curve TED\_ML is near linear, which might be a side-effect of sentence-length normalization. 
CDS produces sentences that better match the average sentence length in the in-domain corpus, which was also observed in \citet{santamaria2017data}.\\

\begin{figure}
    \centering
    \includegraphics[width=\linewidth]{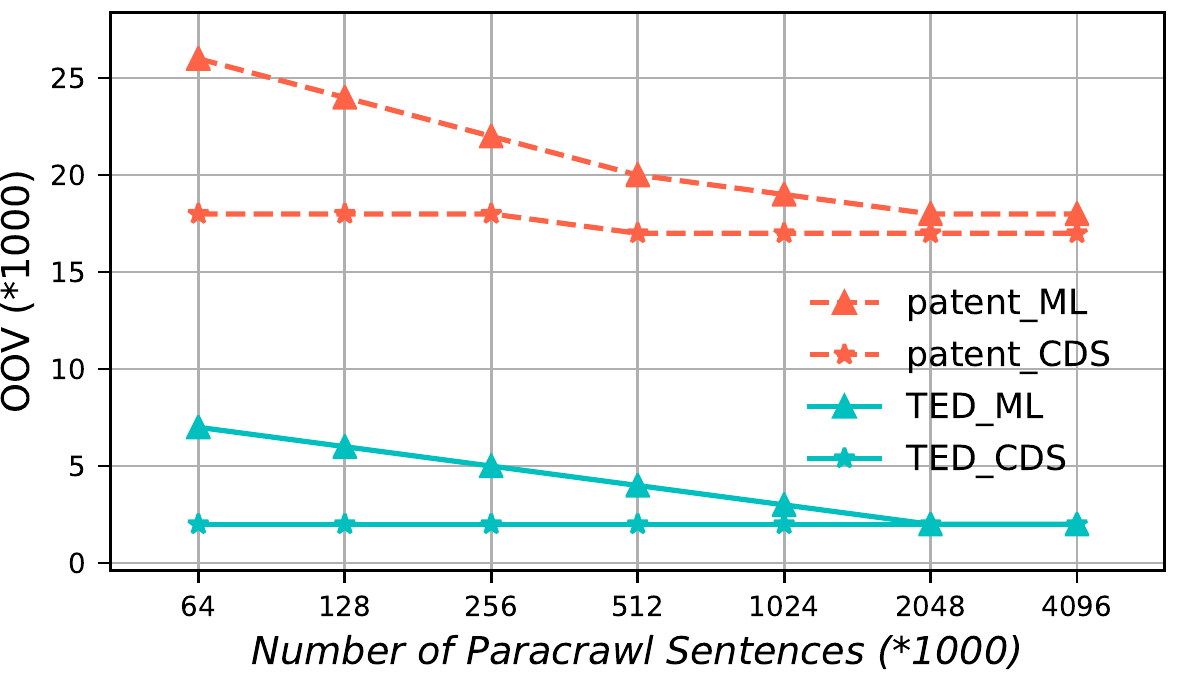}
    \caption{Number of OOV words in the source side of German-English target domain corpora.}
    \label{fig:oov}
\end{figure}

\noindent{\bf Out-of-Vocabulary Words}  We count out-of-vocabulary (OOV) tokens in in-domain corpus based on the vocabulary of selected unlabeled-domain data (Figure \ref{fig:oov}). The CDS subsets cover in-domain vocabulary better than ML subsets as expected, since CDS is based on vocabulary coverage.\\

\begin{figure}
    \centering
    \includegraphics[width=\linewidth]{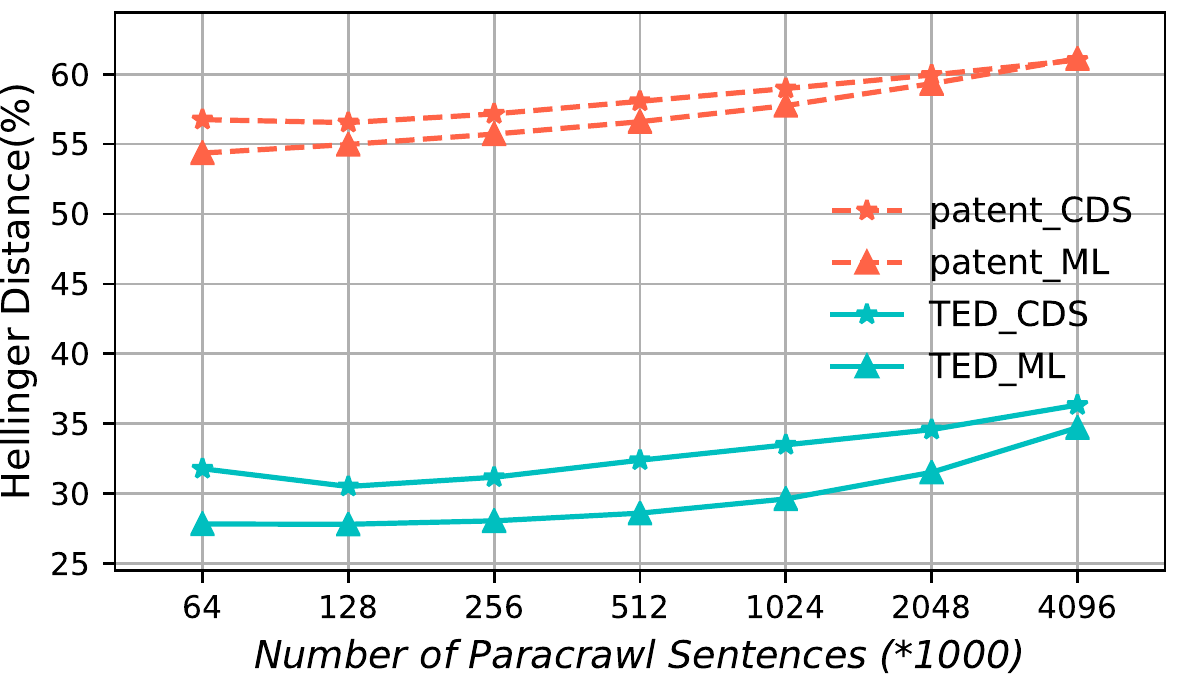}
    \caption{Hellinger distance for source side unigram distributions of German-English corpora between in-domain data and ML/CDS selected data. }
    \label{fig:hellinger}
\end{figure}

\noindent{\bf Unigram Distribution Distance} How do unigram relative frequencies compare in the in-domain and selected Paracrawl data?

 We measure the difference of unigram distributions from two corpora by {\em Hellinger distance}, which is defined as Equation \ref{eq:hd} when the probability distribution is discrete, where $P$ and $Q$ are the unigram distributions for the source side of in-domain and Paracrawl. $V$ is the vocabulary size.\footnote{In Figure \ref{fig:hellinger}, for the purpose of fair comparison, each distribution is defined on the same vocabulary, consisting of the source side vocabulary of TED, patent and Paracrawl data.}
\begin{equation}
    H_{HD}(P,Q) = \frac{1}{\sqrt{2}} \sqrt{\sum^V_{i=1}(\sqrt{p_i}-\sqrt{q_i})^2}.
\label{eq:hd}
\end{equation} 

From Figure~\ref{fig:hellinger}, we can see ML can better match the in-domain vocabulary distribution than CDS. 

With respect to the OOV rate and unigram distribution, patent is more distant from the Paracrawl data than TED is. Figure \ref{fig:results} suggests that CDS dominates ML for distant domains such as Patent, while ML can do slightly better than CDS for domains that are not that distant such as TED.

\section{Related Work}
\label{sec:background}

Curriculum learning has shown its potential to improve sample efficiency for neural models \citep{graves2017automated, weinshall2018curriculum} by guiding the order of presented samples, usually from easier-to-learn samples to difficult samples. Although there is no single criterion to measure difficulty for general neural machine translation tasks
\citep{kocmi2017curriculum, wang2018denoising, zhang2018empirical, kumar2019rl, platanios2019cb}, for the domain adaptation scenario, we measure difficulty based on the distance from in-domain data. Compared to previous work, our application of curriculum learning mainly focuses on improvements on translation quality without consideration of convergence speed. 

\citet{chu2018survey} surveyed recent domain adaptation methods for NMT. In their taxonomy, our workflow in Figure \ref{fig:intro} can be considered a hybrid that uses both data-centric  and model-centric techniques due to the use of additional unlabeled-domain data, with a modified training procedure based for continued training. 

For data-centric domain adaptation methods, our curriculum learning approach has connections to instance weighting. In our work, the presentation of certain examples at specific training phases is equivalent to up-weighting those examples and down-weight the others at that time. Weights of similar samples and less similar ones are adjusted dynamically during the training of NMT models based on the curriculum training strategy. 
In NMT, instance weighting is usually implemented by modifying the objective function \citep{chen2016semi, wang2017instance, chen2017cost}.
In statistical machine translation, \citet{matsoukas2009discriminative} extract features from sentences to capture their domains and then use a classifier to map features to sentence weights. \citet{foster2010discriminative} extend this method by weighting at the level of phrase pairs.
\citet{shah2010translation} use resampling to weight corpora and alignments. \citet{mansour2012simple} focus on sentence-level weighting for phrase extraction. \citet{zhou2015domain} weight examples based on their word distributions. 

For model-centric domain adaptation methods, our work is related to \citet{van2017dynamic}. They adopt gradual fine-tuning, which does the opposite of our method: training starts from the whole dataset, and the training set gradually decreases by removing less similar sentences. \citet{wang2018denoising} use a similar approach, where the NMT model is trained on progressively noise-reduced data batches. However, such schedules have the risk of wasting computation on non-relevant data, especially when most of the Paracrawl data is not similar to the target domain. 


\section{Conclusion}

We introduced a curriculum learning approach to adapt neural machine translation models to new domains. Our approach first ranks unlabeled-domain training samples based on their similarity to in-domain data, and then adopts a probabilistic curriculum learning strategy so that more similar samples are used earlier and more frequently during training. 

We show the effectiveness of our method on four tasks. Results show that curriculum learning models can improve over the standard continued training model by up to 3.22 BLEU points and can take better advantage of distant and noisy data. According to our S$^4$ analysis of lexical choice errors, this improvement is mainly due to better scoring of words that acquire a new SENSE or have a different SCORE distribution in the new domain.  
Our extensive empirical analysis suggests that this approach is effective for several reasons: 
(1) It provides a robust way to augment the training data with samples that have different levels of similarity to the in-domain data. Unlabeled-domain data such as webcrawls naturally have a diverse set of sentences, and the probabilistic curriculum allows us to exploit as much diversity as possible.
(2) It implements the intuition that samples more similar to in-domain data are seen earlier and more frequently; when adding a new shard into the training set, the previously visited shards are still used, so the model will not forget what it just learned.
(3) It builds on a strong continued training baseline, which continues on in-domain data. (4) The method implements best practices that have shown to be helpful in NMT, e.g.~bucketing, mini-batching, and data shuffling. 

For future work, it would be interesting to measure how curriculum learning models perform on the general domain test set (rather than the in-domain test set we focus on in this work); do they suffer more or less from catastrophic forgetting \citep{goodfellow2013empirical, kirkpatrick2017overcoming, Khayrallah2018regularized, Thompson2019NAACL}?


\section*{Acknowledgments}
This work is supported in part by a AWS Machine Learning Research Award and a grant from the Office of the Director of National Intelligence, Intelligence Advanced Research Projects Activity (IARPA), via contract \#FA8650-17-C-9115. The views and conclusions contained herein are those of the authors and should not be interpreted as necessarily representing the official policies, either expressed or implied, of the sponsors. 

We thank the organizers and participants of the 2018 Machine Translation Marathon for providing a productive environment to start this project. We also thank Amittai Axelrod, Hongyuan Mei and all the team members of the JHU SCALE 2018 workshop for helpful discussions.


\bibliography{reference}
\bibliographystyle{acl_natbib}

\clearpage
\appendix
\section{In-domain Data Details}
\label{app:indata}
\begin{table}[h]
    \centering
    \small
    \begin{tabular}{p{0.9cm} c c c c}
         \hline
         {\bf Dataset} &  TED(de) & TED(ru) & patent(de) & patent(ru)\\
         \hline
         {\bf \#samples} & 151,627 & 180,316 & 821,267 & 28,536\\
         \hline
    \end{tabular}
    \caption{In-domain data statistics.}
    \label{tab:indata}
\end{table}
The total amount of the in-domain data in each domain is summarized in Table \ref{tab:indata}. In this paper, we uniformly sample 15k in-domain data from each dataset. We choose the amount of 15k, which makes up a relatively small percentage of the original corpora, in order to evaluate the extreme case of low-resource domain adaptation settings. Under this setting, the positive effect of adding more selected unlabeled-domain data into training corpus is more obvious in terms of the performance improvement of NMT models. Our pilot experiments show that curriculum learning can scale with more in-domain data---it consistently outperforms the standard training policy, but with less improvement. This is not surprising, as when there is enough in-domain data, continued training on only the in-domain data can already achieve a pretty good performance, and we do not need to use  extra unlabeled-domain data to augment it any more, neither does curriculum learning.

\section{Data Sharding}
\label{app:dsharding}
\begin{figure}[h]
    \centering
    \includegraphics[width=\linewidth]{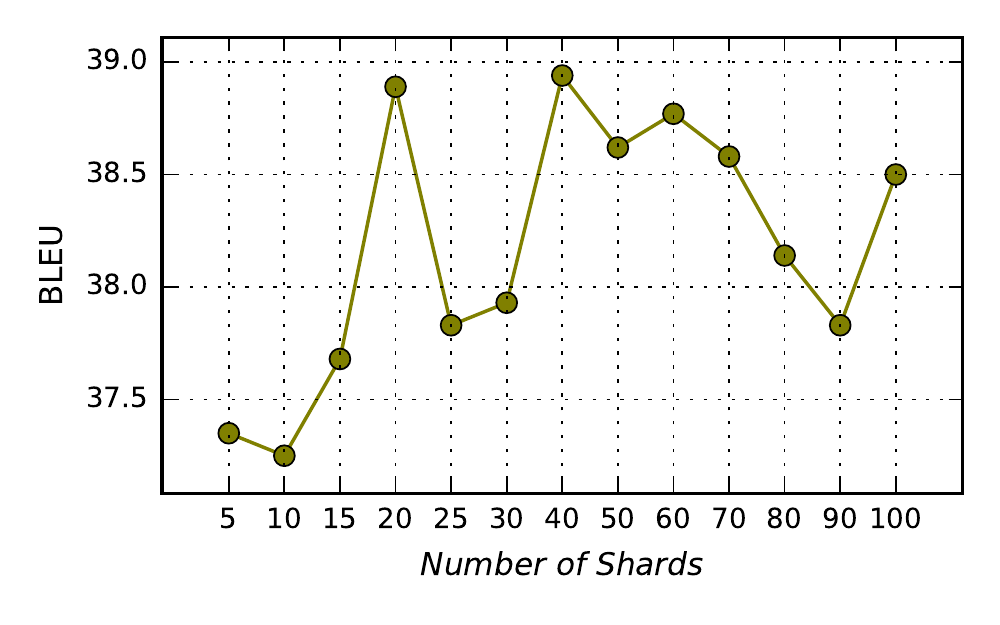}
    \caption{Tuning number of shards on a curriculum learning model trained with German-English corpus augmented by 1024k Paracrawl data.}
    \label{fig:shards}
\end{figure}

We experimented with different number of shards for curriculum learning models as shown in Figure \ref{fig:shards}. Overall, the performance shows the tendency to first improve and then degrade as the number of shards increases. 
Consider the extreme case, where the data are all put into one shard, or there are as many shards as samples, then it will actually be the same as the standard continued training. 

\section{S$^4$ Error Analysis Algorithm}
\label{app:s4}
The algorithm for getting S$^4$ word translation error statistics is shown in Algorithm \ref{alg:s4}.
\begin{algorithm}[h]
	\begin{algorithmic}[1]
	\LeftComment{$S$: The source-side sentences in test set}
	\LeftComment{$f_i$: The $i$th unique word in a sentence}
	\LeftComment{$E^r(f_i)$: Words aligned to $f_i$ in the reference translation for test set}
	\LeftComment{$E^h(f_i)$: Words aligned to $f_i$ in the output translation for test set}
	\LeftComment{$E^t(f_i)$: Words aligned to $f_i$ in the reference translation for training set}
	\Procedure{S4ErrorCounter}{$S,E^r,E^h,E^t$}
	\State correct$\gets$0; seen$\gets$0; sense$\gets$0; score$\gets$0
	\For{$s \in S$}
	\For{$f_i \in s$}
	\For{$e_j \in E^r(f_i)$}
	    \If{$e_j \in E^h(f_i)$}
	        \State correct $\gets$ correct+1
	    \Else 
	        \If{$f_i \notin E^t$}
	            \State seen $\gets$ seen+1
	        \Else 
	            \If{$e_j \notin E^t(f_i)$}
	                \State sense $\gets$ sense+1
	            \Else
	                \State score $\gets$ score+1
	            \EndIf
	         \EndIf
	     \EndIf
	 \EndFor
	 \EndFor
	 \EndFor
	 \State {\bf return} correct, seen, sense, score
	 \EndProcedure
	
	\end{algorithmic}
	\caption{S$^4$ Error Analysis}
	\label{alg:s4}
\end{algorithm}

\section{Bilingual Criterion} 
\label{app:bilingual}
\begin{figure}[h]
    \centering
    \includegraphics[width=\linewidth]{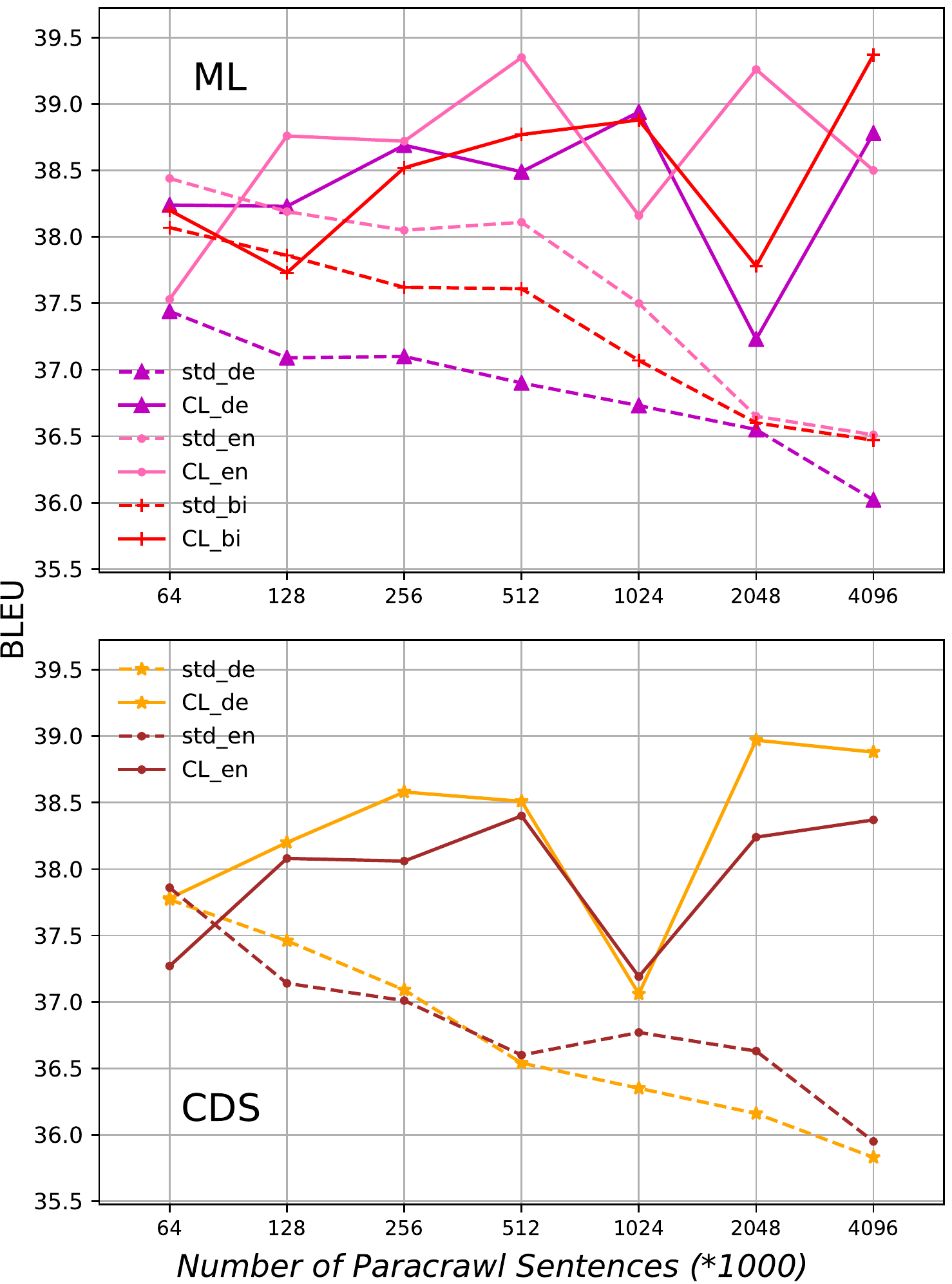}
    \caption{Performance of models continued trained with TED data and Paracrawl data, ranked by their similarity scores collected from the source side (de) or the target side (en) or both sides (bi) of the sentence pairs.}
    \label{fig:bilingual}
\end{figure}

In previous experiments, we only considered the data selection scores obtained from the source side of the corpora. It is very likely that curriculum learning would benefit from also taking into account the features of target side. For Moore-Lewis score, following \citet{axelrod2011domain}, we sum the scores over each side of the corpus:
\begin{equation}
    [H_{I-src}(s)-H_{N-src}(s)] + [H_{I-tr}(s)-H_{N-trg}(s)].
\end{equation}
In addition, we also conduct comparison experiments using scores obtained from only the target side for both of the two data selection methods. Results are shown in Figure \ref{fig:bilingual}.

For Moore-Lewis method, in terms of the standard models, scores collected from target side can lead to better translation than source side scores, and bilingual criteria is somewhere in between, for all the sizes of Paracrawl data we experimented with. But this does not map to the curriculum learning models perfectly. Although CL\_en achieves several impressive BLEU scores (39.35 BLEU at 512k, 39.26 BLEU at 2048k), CL\_de can sometimes outperform it. And the performance of their bilingual counterparts are unpredictable: it can be either worse or better than both of them. At 4096k ML selected sentences, CL\_bi improves the BLEU score to 39.37 BLEU, which is the best test score among all the results we have for German-English TED. For cynical data selection, it is obvious that curriculum learning models prefer the scores obtained from the source side of the corpus. 

\section{Perplexity Selection} 
\begin{figure}[h]
    \centering
    \includegraphics[width=\linewidth]{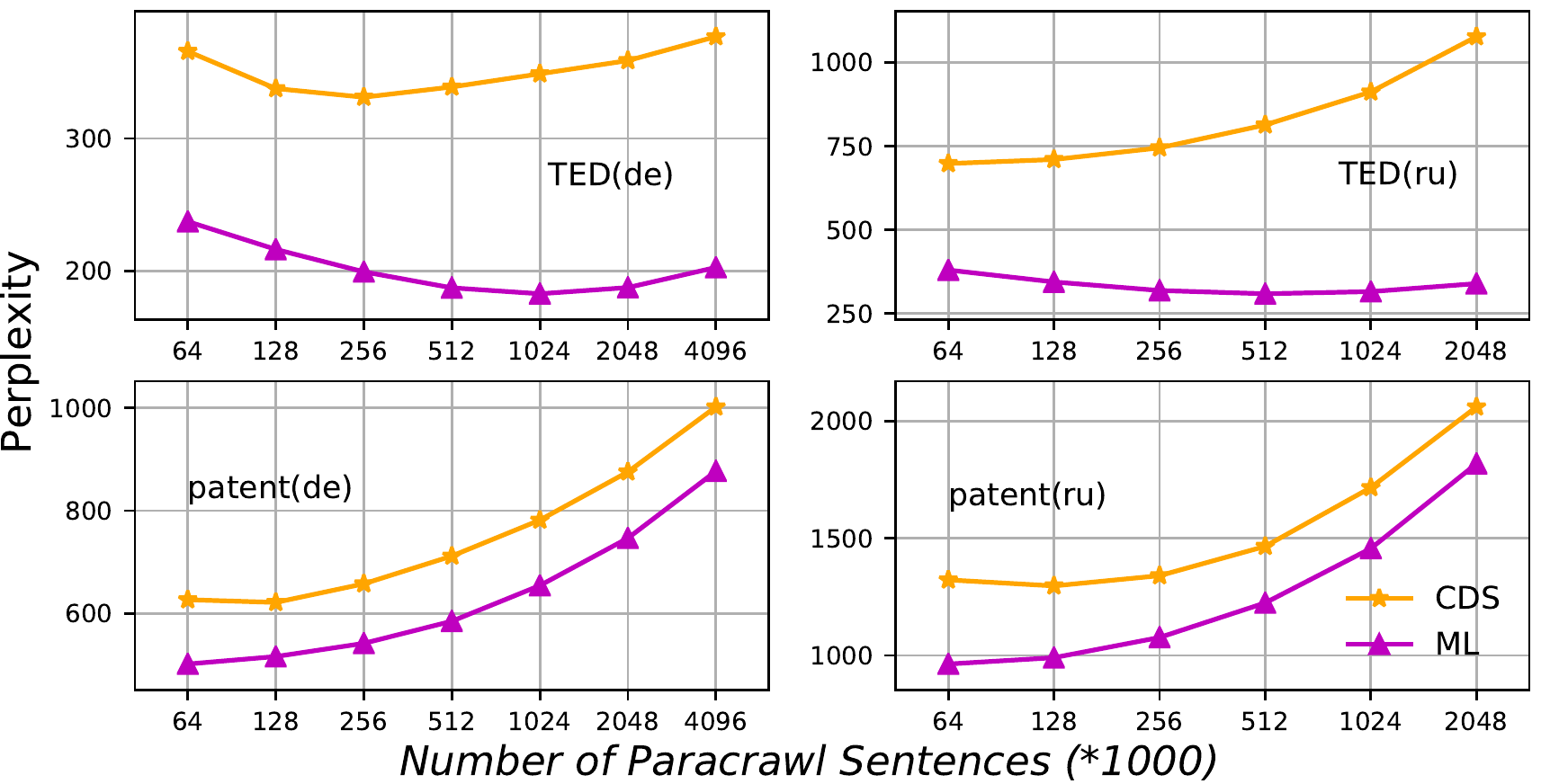}
    \caption{Perplexity of selected data evaluated on the language model learned from in-domain corpus.\\}
    \label{fig:ppl}
\end{figure}
In Section \ref{ssec:results}, we train models with Paracrawl data of different sizes, only after the training is finished and we get the decoding results of those NMT systems, as shown in Figure \ref{fig:results}, can we know which size should be the best choice. \citet{moore2010intelligent} proposed a method that can determine the count cut-offs of the selected data beforehand, so that a lot of time and computation will be saved. In their work, the optimal selection threshold is determined by the perplexity of in-domain set evaluated on the language models trained on the different-size selected subsets. We name this method as perplexity selection and we are curious whether it is effective in the NMT settings. Unfortunately, the best thresholds elected by this method (Figure \ref{fig:ppl}) are inconsistent with the cutoffs that achieve high BLEU scores in Figure \ref{fig:results}. We can then conclude that perplexity selection may not be an appropriate way to determine the optimal amount of unlabeled-domain data to use for NMT models. 

However, if computational resources are limited, according to the experiment results (Figure \ref{fig:results}) in our work, we recommend 1024k as the first choice for cutoffs on ranked unlabeled-domain data, for NMT domain adaptation models trained with curriculum learning strategy.

\end{document}